\documentclass{article}

\usepackage{microtype}
\usepackage{amsmath,nccmath}
\usepackage{booktabs} % for professional tables
\usepackage{xspace,times}
\usepackage{amsmath,amsfonts,bm}

\def\eqref#1{equation~\ref{#1}}
\def\1{\bm{1}}

\DeclareMathAlphabet{\mathsfit}{\encodingdefault}{\sfdefault}{m}{sl}
\SetMathAlphabet{\mathsfit}{bold}{\encodingdefault}{\sfdefault}{bx}{n}

\newcommand{\R}{\mathbb{R}}

\usepackage{hyperref}

\usepackage{url}
\usepackage{graphicx,float}
\usepackage[font=small,labelfont=bf]{caption}
\usepackage{wrapfig,lipsum,booktabs}
\usepackage{subcaption}

\usepackage[accepted]{icml2019}

\usepackage{array,multirow}

\newcommand{\omitme}[1]{}

\usepackage{pgfplots}
\usetikzlibrary{shapes.geometric}
\pgfplotsset{compat=1.7}

\icmltitlerunning{GDPP: Learning Diverse Generations using Determinantal Point Processes}
\begin{document}

\twocolumn[
\icmltitle{GDPP: Learning Diverse Generations using Determinantal Point Processes}

% \author{Mohamed Elfeki$^\dagger$, Camille Couprie$^\ddagger$, Morgane Rivi\`ere$^\ddagger$ \& Mohamed Elhoseiny$^\ddagger$  \\
% $^\dagger$ University of Central Florida, $^\ddagger$ Facebook Artificial Intelligence Research \\
% \texttt{elfeki@cs.ucf.edu, \{coupriec,mriviere,elhoseiny\}@fb.com} \\
% }

\begin{icmlauthorlist}
\icmlauthor{Mohamed Elfeki}{ucf}
\icmlauthor{Camille Couprie}{fb}
\icmlauthor{Morgane Rivi\`ere}{fb}
\icmlauthor{Mohamed Elhoseiny}{fb,kaust}
\end{icmlauthorlist}
% \begin{center}
% $^1$ University of Central Florida, $^2$ Facebook Artificial Intelligence Research \\  $^3$ King Abdullah University of Science and Technology\\
% \texttt{elfeki@cs.ucf.edu, \{coupriec,mriviere\}@fb.com, mohamed.elhoseiny@kaust.edu.sa} \\
% \end{center}

\icmlaffiliation{ucf}{University of Central Florida}
\icmlaffiliation{fb}{Facebook Artificial Intelligence Research}
\icmlaffiliation{kaust}{King Abdullah University of Science and Technology}

\icmlcorrespondingauthor{Mohamed Elfeki}{elfeki@cs.ucf.edu}
\icmlcorrespondingauthor{Couprie,Rivi\`ere}{\{coupriec,mriviere\}@fb.com} $\,\,\,\,\,\,\,\,\,\,\,\,\,\,$
\icmlcorrespondingauthor{Mohamed Elhoseiny}{mohamed.elhoseiny@kaust.edu.sa}
\vskip 0.3in
]
\printAffiliationsAndNotice{}

\begin{abstract}
Generative models have proven to be an outstanding tool for representing high-dimensional probability distributions and generating realistic looking images. An essential characteristic of generative models is their ability to produce multi-modal outputs. However, while training, they are often susceptible to mode collapse, that is models are limited in mapping input noise to only a few modes of the true data distribution. In this work, we draw inspiration from Determinantal Point Process (DPP) to propose an unsupervised penalty loss that alleviates mode collapse while producing higher quality samples. DPP is an elegant probabilistic measure used to model negative correlations within a subset and hence quantify its diversity. We use DPP kernel to model the diversity in real data as well as in synthetic data. Then, we devise an objective term that encourages generator to synthesize data with a similar diversity to real data. In contrast to previous state-of-the-art generative models that tend to use additional trainable parameters or complex training paradigms, our method does not change the original training scheme. Embedded in an adversarial training and variational autoencoder, our Generative DPP approach shows a consistent resistance to mode-collapse on a wide-variety of synthetic data and natural image datasets including MNIST, CIFAR10, and CelebA, while outperforming state-of-the-art methods for data-efficiency, generation quality, and convergence-time whereas being 5.8x faster than its closest competitor. \footnote{https://github.com/M-Elfeki/GDPP}

% Our code, attached to the submission, will be made publicly available. 

% Moreover, GDPP only adds 10\% running time overhead to the default DCGAN, as opposed to state-of-the-art that adds more than 500\% overhead. 

%\mohamed{state some key results here, for example that our overhead is 25\% vs 500\% for WGAN-GP, this makes the key value more clear. Also say something about quality and coverage. We cover XXX\% of modes with YYY quality, }\elfeki{I added concrete results to the contribution subsection in the Intro, because the abstract is quite long as you said.}

\end{abstract}

\section{Introduction}
Deep generative models have gained great research interest in recent years as a powerful framework to represent high dimensional data in an unsupervised fashion. Among many generative approaches, Generative Adversarial Networks (GANs)~\cite{goodfellow2014generative} and Variational AutoEncoders (VAEs)~\cite{vae} took a place among the most prominent approaches for synthesizing realistic images. They consist of two networks: a generator (decoder) and a discriminator (encoder), where the generator attempts to map latent code to \textit{fake} data points that simulate the distribution of \textit{real} data. Nevertheless, in the process of learning multi-modal complex distributions, both models may converge to a trivial solution where the generator learns to produce few modes exclusively, which referred to by mode collapse. 
 
\begin{figure}
    \centering
    \includegraphics[width=0.4\textwidth,height=4.8cm]{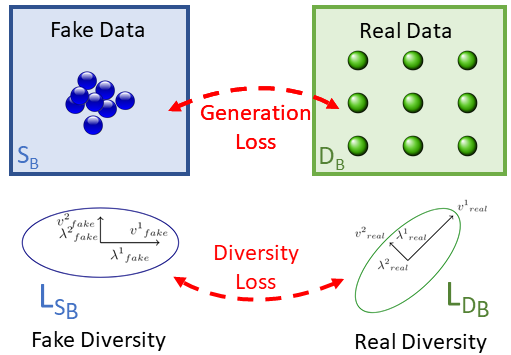}
    \vspace{-1em}
    \captionof{figure}{Inspired by DPP, we model a batch diversity using a kernel $L$. Our loss encourages generator $G$ to synthesize a batch $S_B$ of a diversity $L_{S_B}$ similar to the real data diversity $L_{D_B}$, by matching their eigenvalues and eigenvectors. Generation loss aims at generating similar data points to the real, and diversity loss aims at matching the diversity manifold structures. }
    \label{fig:diversity_loss}
    \vspace{-1em}
\end{figure}
% \mohamed{I think the teaser figures could be simpler and as interesting as possible and more intuitive. I am thinking of something that brings.}\elfeki{ ==> Like what?} \mohamed{I liked the ellipse illustration, remember that we had that at some point}
% \mohamed{We propose to model generative diversity probabilistically with Determinantal Point Processes (DPP)}

To address this, we propose using Determinantal Point Processes (DPP) to model the diversity within data samples. DPP is a probabilistic model that has been mainly adopted for solving subset selection problems with diversity constraints~\citep{learning_dpp}, such as video and document summarization. In such cases, representative sampling requires quantifying the diversity of $2^N$ subsets, where $N$ is the size of the ground set. However, this renders DPP sampling from true data to be computationally inefficient in the generation domain. The key idea of our work is to model the diversity within real and fake data throughout the training process using DPP kernels, which adds an insignificant computational overhead. Then, we encourage producing samples of similar diversity distribution to the true data by back-propagating our proposed DPP-inspired metric through the generator. In such a way, the generator explicitly learns to cover more modes of real distribution without a significant overhead.

Recent approaches tackled mode-collapse in one of two different ways: (1) modifying the learning of the system to reach a better convergence point (e.g. \cite{unrolled_gan,Gulrajani2017improved}); or (2) explicitly enforcing the models to capture diverse modes or map back to the true-data distribution (e.g. \cite{veegan,mode_gan}). Here we focus on a relaxed version of the latter, where we use the same learning paradigm of the standard generators and add a penalty term to the objective function. The advantage of such an approach is to avoid adding any extra trainable parameters to the framework while maintaining the same back-propagation steps as the default learning paradigm. Thus, our model converges faster to a fair equilibrium point where generator imitates the diversity of true-data distribution and produces higher quality generations. 

\textbf{Contribution}. we introduce a new penalty term, that we denote {Generative Determinantal Point Processes (\textit{GDPP}) loss}. Our loss only assumes access to a generator $G$ and a feature extraction function $\phi(\cdot)$. The loss encourages the generator to diversify generated samples to match the diversity of real data as illustrated in Fig.~\ref{fig:diversity_loss}. This criterion can be considered as a complement to the original generation loss which attempts to learn an indistinguishable distribution from the true-data distribution without explicitly enforcing diversity. We assess the performance of GDPP on three different synthetic data environments, while also verifying its advantage on three real-world images datasets. Our approach consistently outperforms several state-of-the-art approaches that of more complex learning paradigms in terms of alleviating mode-collapse and generation quality. 

\section{Related Work}
Among many existing generation frameworks, GANs tend to synthesize the highest quality generations, however, they are harder to optimize due to unstable training dynamics. Here, we discuss a few generic approaches addressing mode collapse with an emphasis on GANs. We categorize them based on their approaches to alleviate mode collapse.

 %\mohamed{I think related work needs to be organized more around key ideas and contrast against our work.}
 %\mohamed{why did you organize the related work in this partyicular way and why this way of presenting it may best related to DPP. I liked the local development of the sentences. My concerns is more global (the tip of thee ice cream cone in writing). I think it is nice to say a statement or two here to introduce at a high levels this organization. We may think how to better improve that. Should for example bayesian modeling be a different package. } 
%\elfeki{You mean to have another paragraph about Bayesian Modeling?}
%\mohamed{I mean, it will be good to introduce why you are presenting the literature in that way beforehand, so that the global taste of the section is clear before hand}\elfeki{I've modified the first paragraph to provide an overview on this section.}

\vspace{-1em}
\paragraph{Mapping generated data back to noise.} \citep{bigan,ALI} are of the earliest methods that proposed learning a reconstruction network besides learning the generative network. Adding this extra network to the framework aims at reversing the action of generator by mapping from data to noise. Likelihood-free variational inference (LFVI) \citep{lfvi}, merges this concept with learning implicit densities using hierarchical Bayesian modeling. Ultimately, VEEGAN \citep{veegan} used the same concept, but without basing reconstruction loss on the discriminator. This has the advantage of isolating the generation process from the discriminator's sensitivity to any of the modes. Along similar lines, \cite{mode_gan} proposed several ways of regularizing the objective of adversarial learning including geometric metric regularizer, mode regularizer, and manifold-diffusion training. Specifically, mode regularization has shown a potential into alleviating mode collapse and stabilizing the training. 
\vspace{-1em}
\paragraph{Providing a surrogate objective function.}
InfoGAN \cite{infogan} propose an information-theoretic extension of GANs that obtains disentangled representation of data by latent-code reconstitution through a penalty term in its objective. InfoGAN includes autoencoder over latent codes; however, it was shown to have stability problems similar to the standard GAN and requires stabilization empirical tricks. The Unrolled-GAN of \cite{unrolled_gan} propose a novel objective to update the generator with respect to the unrolled optimization of the discriminator. This allows training to be adjusted between using the optimal discriminator in the generator's objective, which has been shown to improve the generator training process and to reduce mode collapse. Generalized LS-GAN of \cite{edraki2018generalized} define a pullback operator to map generated samples to the data manifold. With a similar philosophy, BourGAN \cite{xiao2018bourgan} draws samples from a mixture of Gaussians instead of a single Gaussian. There is, however, no specific enforcement to diversify samples. Finally, improving Wasserstein GANs of \cite{wgan}, WGAN-GP \citep{Gulrajani2017improved} introduce a gradient penalization employed in state-of-the-art systems \citep{karras2017progressive}.

\vspace{-1em}
\paragraph{Using multiple generators and discriminators.}
 One of the popular methods to reduce mode collapse is using multiple generator networks to provide better coverage of the true data distribution. \cite{coupled_gan} propose using two generators with shared parameters to learn the joint data distribution. The two generators are trained independently on two domains to ensure a diverse generation. However, sharing the parameters guide both the generators to a similar subspace. \cite{durugkar} propose a similar idea of multiple discriminators that are being an ensemble, which was shown to produce better quality samples. Recently, \cite{madgan} proposed MAD-GAN which is a multi-agent GAN architecture incorporating multiple generators and one discriminator. Along with distinguishing real from fake samples, the discriminator also learns to identify the generator that synthesized the fake sample. The learning of such a system implies forcing different generators to learn unique modes, which helps in better coverage of data modes. DualGAN of \cite{nguyen2017dual} improves the diversity within GANs at the additional requirement of training two discriminators. The Mixed GAN approach of \cite{lucas2018mixed} rather introduces a permutation invariant architecture for the discriminator, that doubles the number of parameters. In contrast to these approaches, our GDPP-GAN does not require any extra trainable parameters which results in a faster training as well as being less susceptible to overfitting. 
 
 Finally, we also refer to PacGAN \citep{Lin2017PacGANstackedMNIST} which modifies the discriminator input with concatenated samples to better sample the diversity within real data. Nevertheless, such an approach is subject to memory and computational constraints as a result of the significant increase in batch size. Additionally, spectral normalization strategies have been recently proposed in \cite{Miyato2018SNGAN} and SAGAN \citep{Zhang2018SAGAN} to further stabilize the training. We note that these strategies are orthogonal to our contribution and could be implemented in conjunction with ours to further improve the training stability of generative models.

 %%%%%%%%%%%%%%%%%%%%%%%%%%%%%%% Comments %%%%%%%%%%%%%%%%%%%%%%%%%%%%%%%%%

% A large body of work was proposed to tackle the problem of mode collapse in generative models\mohamed{ do we need this sentence? less is more if not needed}. ==>Elfeki: DONE
% \mohamed{Among the existing generative models, GAN has been features with high quality generation, yet, it is harder to optimize} ==>Elfeki: DONE

% \ccc{\cite{madgan} base the ModeGAN method on the assumption of the availability of sufficient samples from every mode in training data. In particular, if a sample from the true data distribution belongs to a particular mode, then the generated fake sample is likely to belong to the same mode with the same probability. Camille: suggesting to remove the above paragraph as the same approach is cited in the next one.} 

 %%%%%%%%%%%%%%%%%%%%%%%%%%%%%%%%%%%%%%%%%%%%% ICLR Submission %%%%%%%%%%%%%%%%%%%%%%%%%%%%%%%%%%%%%%%%%%%%%%%%%%%

%The discriminator is designed to find the real and fake

% \ccc{Finally, we may cite 
% %\ccc{\paragraph{Mini-batch exploitation strategies.}
% PacGAN \cite{Lin2017PacGANstackedMNIST} as another related approach to address mode collapse, by modifying the  discriminator input with concatenated samples. This creates an additional memory and computational overhead.  
% %The Mixed GAN approach of \cite{lucas2018mixed} rather 
% we can cite it later.
% }

\section{Determinantal Point Process (DPP)}
\label{sec:bckg}
\label{sec_dpp}

DPP is a probabilistic measure was introduced in quantum physics~\citep{dpp_quantum} to model the Gauss-Poisson and the 'fermion' processes, then was extensively studied in random matrix theory, e.g.~\citep{dpp_statistics}. It provides a tractable and efficient means to capture negative correlation with respect to a similarity measure, that in turn can be used to quantify the diversity within a subset. As pointed out by \cite{vid_summr_2}, DPP is agnostic about the order of the items within subsets. Hence, it can be used to model data that is randomly sampled from a certain distribution such as mini-batches sampled from training data.

A point process $\mathcal{P}$ on a ground set $\mathcal{V}$ is a probability measure on the power set $2^{N}$, where $N=|\mathcal{V}|$ is the size of the ground set. A point process $\mathcal{P}$ is called determinantal if, given a random subset $Y$ drawn according to $\mathcal{P}$, we have for every $S \subseteq Y$, 
\begin{equation}
\mathcal{P}(S \subseteq Y) \propto \det(L_S)
\label{eq_S}
\end{equation}
for some symmetric similarity kernel $L \in \mathbb{R}^{N\times N}$, where $L_S$ is the similarity kernel of subset $S$. $L$ must be real, positive semidefinite matrix $L \preceq I$ (all the eigenvalues of $L$ are between 0 and 1); since it represents a probabilistic measure and all of its principal minors must be non-negative. 

$L$ is often referred to as the marginal kernel because it contains all the information needed to compute the probability of any subset $S$ being selected in $\mathcal{V}$. $L_S$ denotes the sub-matrix of $L$ indexed by $S$, specifically, $L_S \equiv [L_{ij}];  i,j\in S$. Hence, the marginal probability of including one element $e_i$ is $p(e_i \in Y) = L_{ii}$, and two elements $e_i$ and $e_j$ is $L_{ii} L_{jj} -L_{ij}^2 = p(e_i \in Y) p(e_j \in Y) - L_{ij}^2$. A large value of $L_{ij}$ reduces the likelihood of both elements to appear together in a diverse subset. 

\cite{structured_dpp} proposed decomposing the kernel $L_S$ as a Gram matrix:
\begin{equation}
    % L_{ij} = .
    \mathcal{P}(S \subseteq Y) \propto \det(\phi(S)^\top \phi(S)) \prod_{e_i \in S} q^2(e_i),
    \label{eq:dpp_decomposition}
\end{equation}
where $q(e_i) \geq 0$ can be seen as a quality score of an item $e_i$ in the ground set $\mathcal{V}$, while $\phi_i \in \R^D; D \leq N$ and $||\phi_i||_2 = 1$ is used as an $\ell_2$ normalized feature vector of an item. In this manner, $\phi_i^\top \phi_j \in [-1, 1]$ is evaluated as a "normalized similarity" between items $e_i$ and $e_j$ of $\mathcal{V}$, and the kernel $L_S$ is guaranteed to be real positive semidefinite matrix.

\vspace{-1em}
 \paragraph{Geometric interpretation:} $\det(\phi(S)^\top \phi(S)) = \prod_{i} \lambda_i$, where $\lambda_i$ is the $i^{th}$ eigen value of the kernel $\phi(S)^\top \phi(S)$, and $\lambda \geq 0$ since the kernel is a positive semidefinite matrix. Hence, we may visualize that DPP models diverse representations of data because the determinant of $\phi(S)^\top \phi(S)$ corresponds to the volume in $N$-D which is equivalent to the multiplication of data variances (i.e., the eigen values). 
 
\textbf{DPP in literature:} DPP has proven to be a valuable tool when tackling diversity enforcement in problems such as document summarization (e.g.,  \cite{learning_dpp,doc_sum_1}), pose estimation (e.g., \cite{dpp_pose}) and video summarization (e.g., \cite{vid_summr_2,lstm_dpp_summarization}). For instance, \cite{summarization_lstm} proposed to learn the two parameters $q, \phi$ in eq.~\ref{eq:dpp_decomposition} to quantify the diversity of the kernel $L_S$ based on spatio-temporal features of the video to perform summarization. Recently,~\cite{hsiao2018creating} proposed to use DPP to automatically create capsule wardrobes, i.e.  assemble a minimal set of items that provide maximal mix-and-match outfits given an inventory of candidate garments.

\section{Generative Determinantal Point Processes}
\begin{figure}
    \centering
    \includegraphics[width=0.45\textwidth, height=5.2cm]{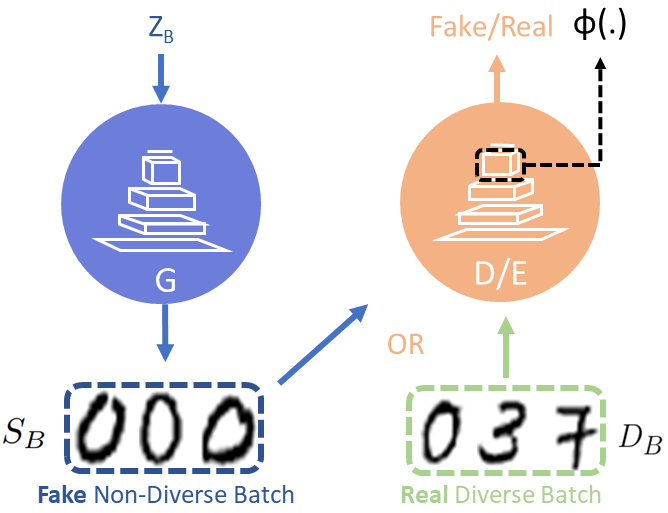}
    \captionof{figure}{Given a generator $G$ and feature extraction function $\phi(\cdot)$, the diversity kernel is constructed as $L=\phi^\top\cdot\phi$. By modeling the diversity of fake and real batches, our loss matches their kernels  $L_{S_B}$ and $L_{D_B}$ to encourage synthesizing samples of similar diversity to true data. We use the last feature map of the discriminator in GAN or the encoder in VAE as the feature representation $\phi$.}
    \label{fig:architecture}
    \vspace{-1em}
\end{figure}

Our GDPP loss encourages the generator to sample fake data of diversity similar to real data diversity. The key challenge is to model the diversity within real data and fake data. We discussed in Sec.~\ref{sec_dpp} how DPP can be used to quantify the diversity within a discrete data distribution. Unlike subset selection problems (e.g., document/video summarization), in the generation domain we are not merely interested in increasing diversity within generated samples. Only increasing the samples diversity will result in samples that are far apart in the generation domain, but not necessarily representative of real data diversity. Instead, we aim to generate samples that imitate the diversity of real data. Thus, we construct a DPP kernel for both the real data and the generated samples at every iteration of the training process as shown in Fig.~\ref{fig:architecture}. Then, we encourage the generator to synthesize samples that have a similar diversity kernel to that of the training data. In order to simplify learning kernels, we match the eigenvalues and eigenvectors of the fake data DPP kernel with their corresponding of the real data DPP kernel. Eigenvalues and vectors capture the manifold structure of both real and fake data, and hence renders the optimization more feasible. Fig.~\ref{fig:diversity_loss} shows pairing the two kernels by matching their high dimensional eigen manifolds.

During training, a generative model $G$ produces a batch of samples $S_B=\{e_1, e_2, \cdots e_B\}; S_B = G(z_B)$, where $B$ is the batch size and $z_B \in R^{d_z \times B}$ is noise vector inputted to the generator $G$. At every iteration, we also have a batch of samples $D_B \sim p_d$, where $p_d$ is a sampler from true distribution. Our aim is to produce $S_B$ that is probabilistically sampled following the DPP kernel of $D_B$, which satisfies:
\begin{equation}
    \mathcal{P}(S_B \subseteq Y) \propto \det(L_{D_B})    
\label{eq_LB}
\end{equation}
such that $Y$ is a random variable representing a fake subset $S_B$ drawn with a generative point process $\mathcal{P}$, and $L_{D_B}$ is DPP kernel of a real subset indexed by $D_B$. 

To construct $L_{S_B}, L_{D_B}$, we use the kernel decomposition in Eq.~\ref{eq:dpp_decomposition}. However, since both true and fake samples are drawn randomly with no quality criteria, it is safe to assume $q(e_i)=1; \forall i\in{1,2,...,B}$. Thus, we construct the kernels as follows: $L_{S_B}=\phi(S_B)^\top \phi(S_B)$ and $L_{D_B}=\phi(D_B)^\top \phi(D_B)$, such that $\phi(S_B)$ and $\phi(D_B)$ are feature representations extracted by the feature extraction function $\phi(\cdot)$.

Our aim is to learn a fake diversity kernel $L_{S_B}$ close to the real diversity kernel $L_{D_B}$. Nonetheless, matching two kernels is an unconstrained optimization problem as pointed out by ~\citep{kernel_unconstrained}. So, instead, we match the kernels using their major characteristics: eigenvalues and eigenvectors. This results in scaling down the matching problem into regressing the magnitudes of eigenvalues and the orientations of eigenvectors. Hence, our devised GDPP loss is composed of two components: diversity magnitude loss $\mathcal{L}_m$, and diversity structure loss $\mathcal{L}_s$ as follows:
\begin{equation}
\begin{aligned}
&\mathcal{L}^{DPP} = \mathcal{L}_m +  \mathcal{L}_s =\\
&\,\,\sum_i \|\lambda^i_{real} - \lambda^i_{fake}\|_2 - \sum_i \hat{\lambda}^i_{real}  \cos(v^i_{real}, v^i_{fake})
\label{eq_GDPP}
\end{aligned}
% \vspace{-0.25em}
\end{equation}
where $\lambda_{fake}^i$ and $\lambda_{real}^i$ are the $i^{th}$ eigenvalues of $L_{D_B}$ and $L_{S_B}$ respectively.

Finally, we account for the outlier structures by using the min-max normalized version of the eigenvalues $\hat{\lambda}^i_{real}$ to scale the cosine similarity between the eigenvectors $v_{fake}^i$ and $v_{real}^i$. This aims to alleviate the effect of noisy structures that intrinsically occur within the real data distribution or within the learning process.

% \vspace{-1em}
\textbf{Integrating GDPP loss with GANs. }  As a primary benchmark, we integrate our GDPP loss with GANs . Since our aim is to avoid adding any extra trainable parameters, we utilize features extracted by the discriminator: we choose to use the hidden activations before the last layer as our feature extraction function $\phi(.)$. We apply $\ell_2$ normalization on the obtained features that guarantees constructing a positive semi-definite matrix according to eq. ~\ref{eq_GDPP}. We finally integrate $\mathcal{L}^{DPP}$ into the GAN objective by only modifying the generator loss of the standard adversarial loss \citep{goodfellow2014generative} as follows:
\begin{equation}
\mathcal{L}_{g} = \mathbb{E}_{z \sim {p_{z}}} [\log(1 -  D(G(z)))] + \mathcal{L}^{DPP}
\end{equation}

% \vspace{-0.5em}
\textbf{Integrating GDPP loss with VAEs. }  A key property of our loss is its generality to any generative model. We show that by also embedding it within VAEs. A VAE consists of an encoder network $q_{\theta_1}(z|x)$, where $x$ is an input training batch and $z$ is sampled from a normal distribution parametrized by encoder outputs $\sigma$ and $\mu$, representing respectively the standard deviation and the mean of the distribution. Additionally, VAE has a decoder network $p_{\theta_2}(x|z)$ which reconstructs $\hat{x}$. We use the final hidden activations in $q$ as our feature extraction function $\phi(.)$. Given a $z$ sampled from a normal distribution $z\sim{\mathcal{N}(\mu,\sigma)}$, $p_{\theta_2}(\hat{x}|z)$ is used to generate the fake batch $S_B$, while the real batch $D_B$ is randomly sampled from training data. Finally, we compute the $\mathcal{L}^{DPP}$ as in Eq.~\ref{eq_GDPP}, rendering the GDPP-VAE loss as:
\begin{equation}
\begin{aligned}
\mathcal{L}_{VAE} = &-\mathbb{E}_{z \sim {q(z|x)}} [\log\{p(x|z)\}]\\ 
&+ KL[q(z|x)||p(z)] + \mathcal{L}^{DPP}.
\end{aligned}
\end{equation}

\begin{table*}[!t]
\centering
\resizebox{\textwidth}{!}{%
{\small
\begin{tabular}{lcccccc}
\hline
& \multicolumn{2}{c}{2D Ring} & \multicolumn{2}{c}{2D Grid}  & \multicolumn{2}{c}{1200D Synthetic}  \\ \hline
 & \begin{tabular}[c]{@{}c@{}}Modes\\ (Max 8)\end{tabular} & \begin{tabular}[c]{@{}c@{}}\% High Quality\\ Samples\end{tabular} & \begin{tabular}[c]{@{}c@{}}Modes\\ (Max 25)\end{tabular} & \begin{tabular}[c]{@{}c@{}}\% High Quality\\ Samples\end{tabular} & \begin{tabular}[c]{@{}c@{}}Modes\\ (Max 10)\end{tabular} & \begin{tabular}[c]{@{}c@{}}\% High Quality\\ Samples\end{tabular} \\ \hline
GAN~\citep{goodfellow2014generative}                                                     & 1                                                       & 99.3                                                              & 3.3                                                      & 0.5                                                               & 1.6                                                      & 2.0                                                               \\
ALI~\citep{ALI}                                                     & 2.8                                                     & 0.13                                                              & 15.8                                                     & 1.6                                                               & 3                                                        & 5.4                                                               \\
Unrolled GAN~\citep{unrolled_gan} & 7.6                                                     & 35.6                                                              & 23.6                                                     & 16.0                                                                & 0                                                        & 0.0                                                               \\
VEE-GAN~\citep{veegan}                                                 & \bf 8.0                                                       & 52.9                                                              & 24.6                                                     & 40.0                                                                & 5.5                                                      & 28.3                                                             \\
WGAN-GP~\citep{Gulrajani2017improved}                                                & 6.8                                                     & 59.6                                                             & 24.2                                                     & 28.7                                                            & 6.4                                                      & 29.5                                                             \\ \hline
GDPP-GAN                                        & \textbf{8.0}                                              & \textbf{71.7}                                                    & \textbf{24.8}                                            & \textbf{68.5}                                                    & \textbf{7.4}                                             & \textbf{48.3}                                                    \\ \hline
\end{tabular}%
}
}
\caption{Degree of mode collapse and sample quality on mixtures of Gaussians. GDPP-GAN consistently
captures the highest number of modes and produces better samples.}
\label{tab:synthetic_table}
\end{table*}
\begin{figure*}[!h]
{\centering
\begin{tabular}{cccccc}
GAN & ALI & Unrolled-GAN & VEE-GAN & WGAN-GP & GDPP-GAN\\
\includegraphics[width=0.13\textwidth]{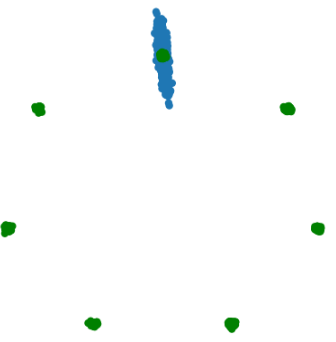}&
\includegraphics[width=0.13\textwidth]{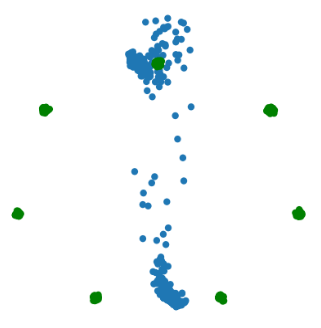}&
\includegraphics[width=0.14\textwidth]{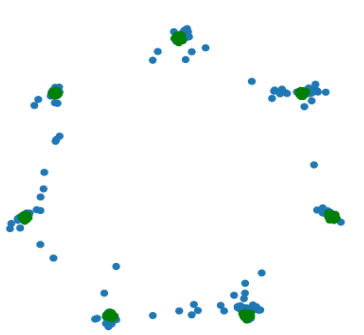}&
\includegraphics[width=0.13\textwidth]{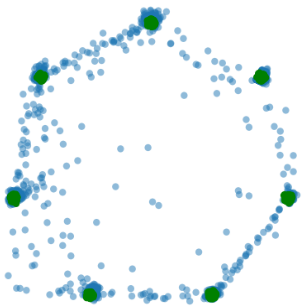}&
\includegraphics[width=0.13\textwidth]{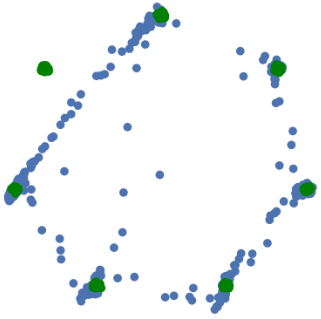}&
\includegraphics[width=0.13\textwidth]{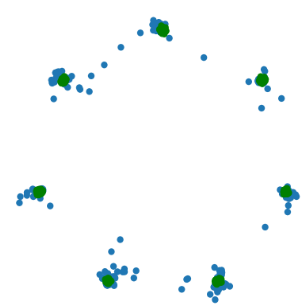}\\
(a) & (b) & (c) & (d) & (e) & (f)\\
\includegraphics[width=0.13\textwidth]{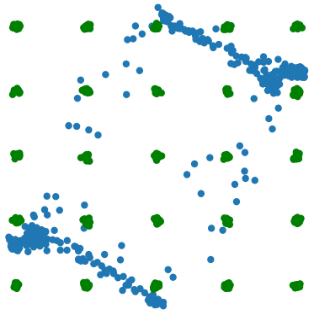}&
\includegraphics[width=0.15\textwidth]{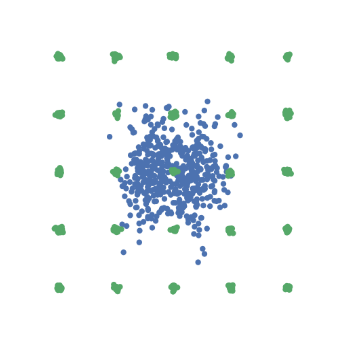}&
\includegraphics[width=0.13\textwidth]{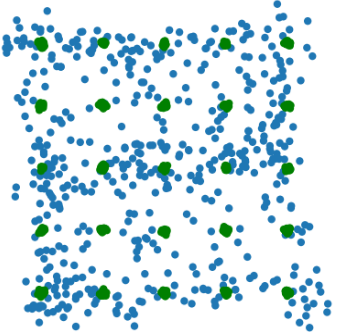}&
\includegraphics[width=0.13\textwidth]{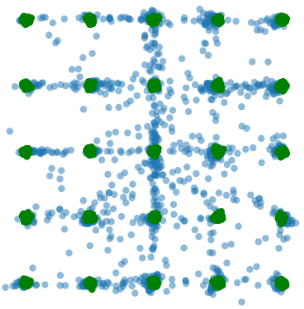}&
\includegraphics[width=0.13\textwidth]{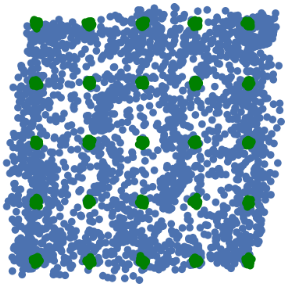}&
\includegraphics[width=0.13\textwidth]{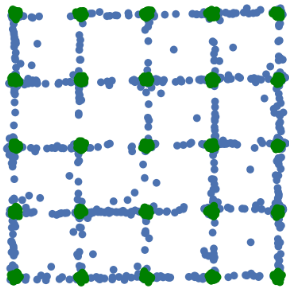}\\
(g) & (h) & (i) & (j) & (k) & (l)\\
\end{tabular}
\captionof{figure}{Scatter plots of the true data (green dots) and generated data (blue dots) from different GAN methods
trained on mixtures of 2D Gaussians arranged in a ring (top) or a grid (bottom). }
\label{fig:synthetic_qualitative}}
\vspace{-3ex}
\end{figure*}

\section{Experiments}
In our experiments, we target evaluating the generation based on two criteria: mode collapse and generated samples quality. Due to the intractability of log-likelihood estimation, this problem is non-trivial for real data. Therefore, we start by analyzing the performance on synthetic data where we can accurately evaluate these criteria. Then, we demonstrate the effectiveness of our method on real data using standard evaluation metrics. The same architecture is used for all methods and hyperparameters were tuned separately for each approach to achieve the best performance (See Appendix A for details). 
% \mohamed{Do we have appendices, or should we refer to supplementary here.}

\subsection{Synthetic Data Experiments}
\label{sec:syn}
Mode collapse and the quality of generations can be explicitly evaluated on synthetic data since the true distribution is well-defined. In this section, we evaluate the performance of the methods on mixtures of Gaussian of known mode locations and distribution (See Appendix B for details). We use the same architecture for all the models, which is the same one used by \cite{unrolled_gan} and \cite{veegan}. We note that the first four rows in Table~\ref{tab:synthetic_table} are obtained from \cite{veegan}, since we are using the same architecture and training paradigm. Fig.~\ref{fig:synthetic_qualitative} illustrates the effect of each method on the 2D Ring and Grid data. As shown by the vanilla-GAN in the 2D Ring example (Fig.~\ref{fig:synthetic_qualitative}a), it can generate the highest quality samples however it only captures a single mode. On the other extreme, the WGAN-GP on the 2D grid (Fig.~\ref{fig:synthetic_qualitative}k) captures almost all modes in the true distribution, but this is only because it generates highly scattered samples that do not precisely depict the true distribution. GDPP-GAN (Fig.~\ref{fig:synthetic_qualitative}f,l) creates a precise representation of the true data distribution reflecting that the method learned an accurate structure manifold.
\begin{table*}[]
\centering
\begin{tabular}{lcccc}
\hline
& \multicolumn{2}{c}{2D Ring}   & \multicolumn{2}{c}{2D Grid}                   \\ \hline
& \begin{tabular}[c]{@{}c@{}}Modes\\ (Max 8)\end{tabular} & \begin{tabular}[c]{@{}c@{}}\% High Quality\\ Samples\end{tabular} & \begin{tabular}[c]{@{}c@{}}Modes\\ (Max 25)\end{tabular} & \begin{tabular}[c]{@{}c@{}}\% High Quality\\ Samples\end{tabular} \\ \hline
Exact determinant ($\det{[L_{S_B}]}$) & \bf 8 & \bf 82.9 & 12.6 & 21.7\\
Only diversity magnitude ($\mathcal{L}_m$) & \bf 8 & 67.0 & 20.4 & 15.9\\ 
Only diversity structure ($\mathcal{L}_s$) & \bf 8 & 65.2 & 18.2 & 35.2 \\ 
GDPP with unnormalized structure term ($\mathcal{L}_m + \mathcal{L}_s^u$) & 7.2 & 81.2  & 20.6& \bf 68.8\\
Final GDPP-loss ($\mathcal{L}_m + \mathcal{L}_s$) & \bf 8 & 71.7 & \bf 24.8 & 68.5 \\
\hline
\end{tabular}
\caption{GDPP loss Ablation study on GAN. $\mathcal{L}^{u}_s$ is  the same as $\mathcal{L}_s$ without min-max eigen value normalization.}
\label{tab:ablation}
\vspace{-1em}
\end{table*}

\begin{figure*}[h]
    \centering
    \begin{minipage}{.48\textwidth}
    \begin{tabular}{cc}
    \includegraphics[width=0.45\textwidth, height=2.5cm]{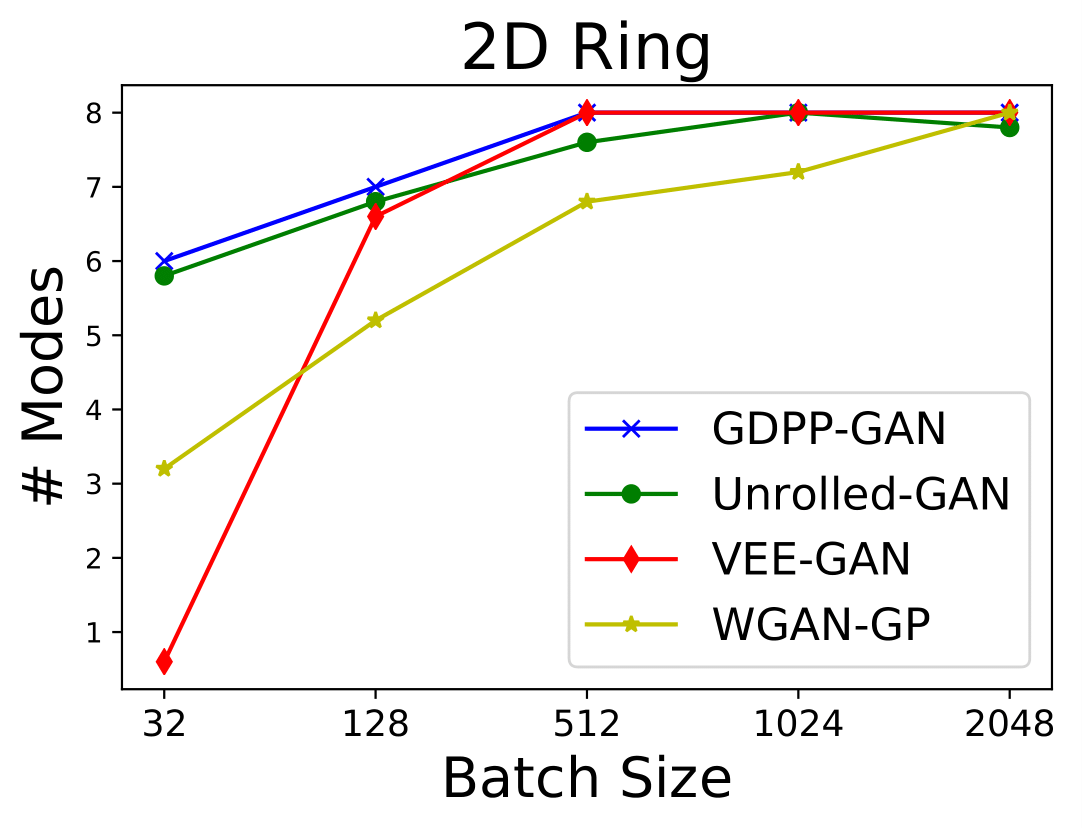}& \includegraphics[width=0.45\textwidth, height=2.5cm]{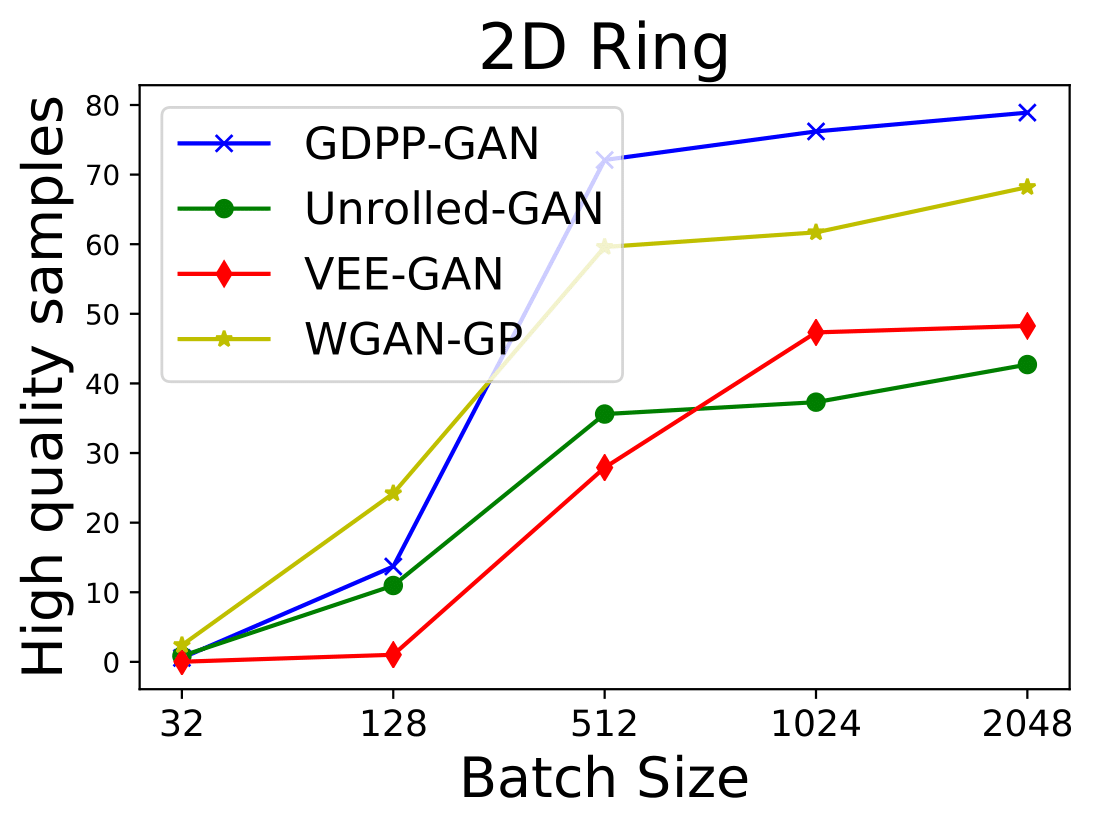}\\
    \includegraphics[width=0.45\textwidth, height=2.5cm]{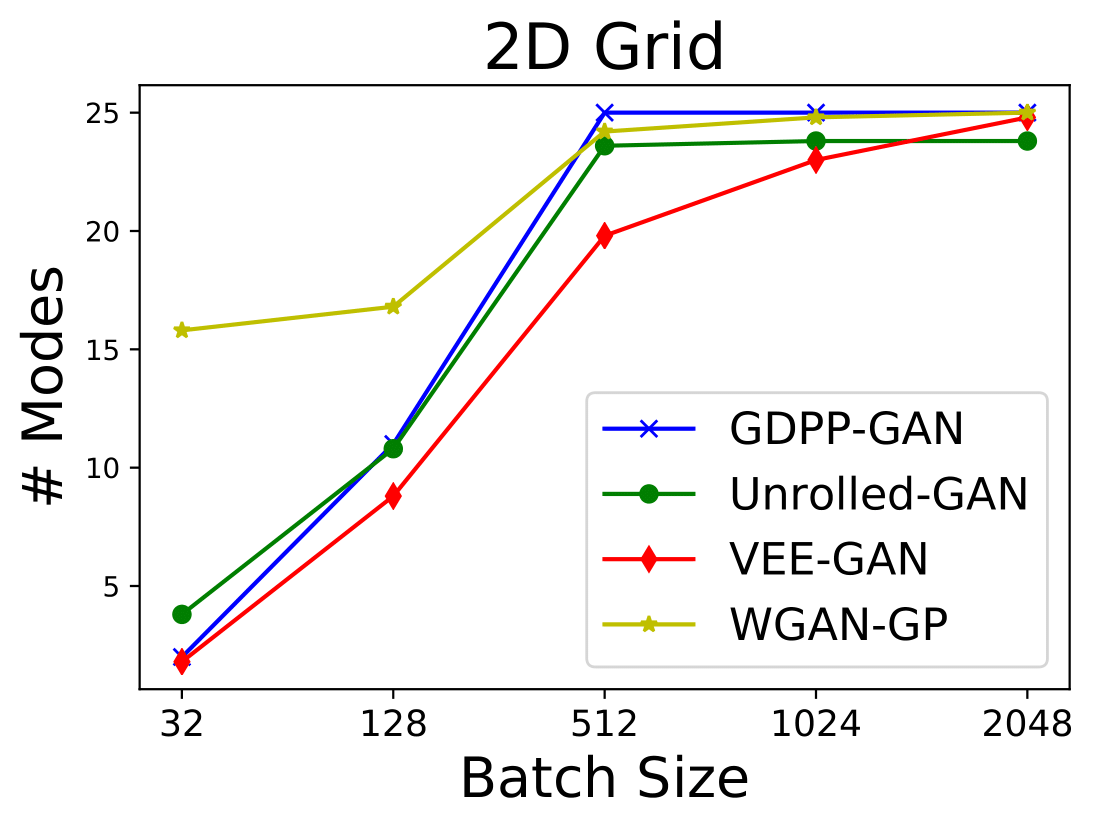}& \includegraphics[width=0.45\textwidth, height=2.5cm]{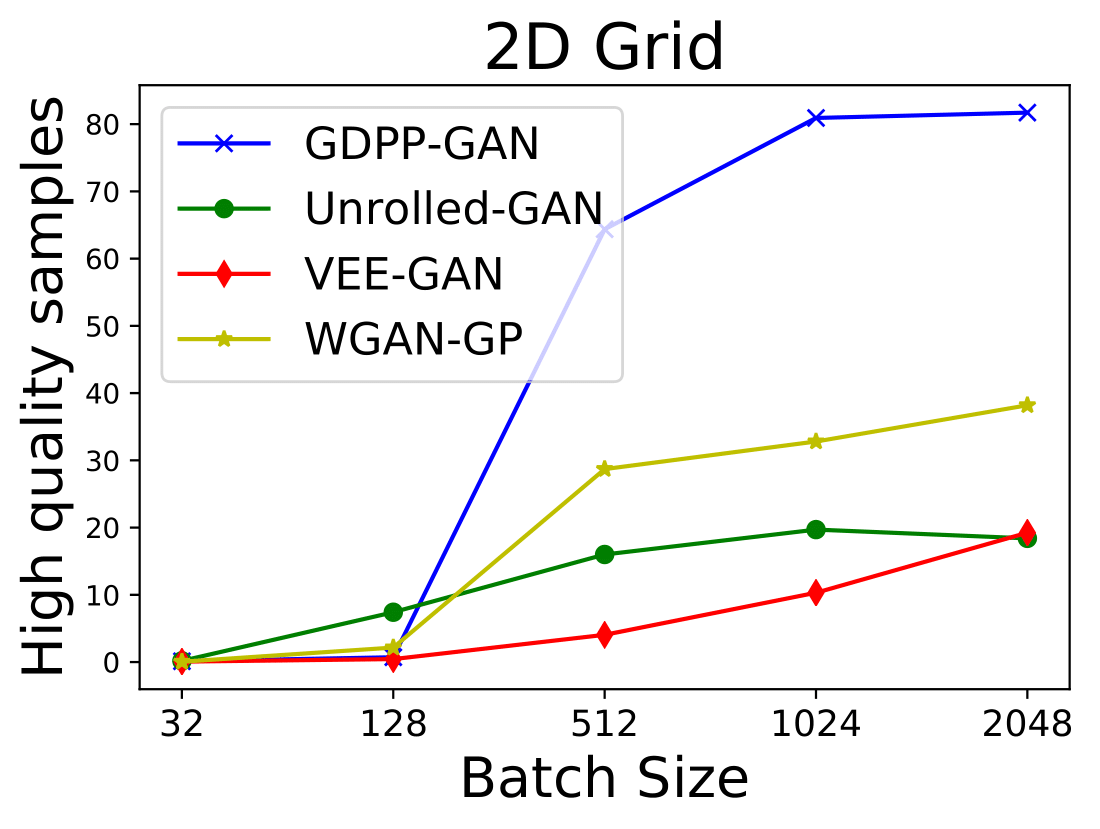}
    \end{tabular}
    \vspace{-1.5em}
    \caption{Data-Efficiency: examining the effect of training batch size $B$ given the same number of training iterations. GDPP-GAN requires the least amount of training data to converge.}
    \label{subfig:batch_size_synth}
    \end{minipage}\hfill
    \begin{minipage}{.48\textwidth}
     \begin{tabular}{cc}
    \includegraphics[width=0.45\textwidth, height=2.5cm]{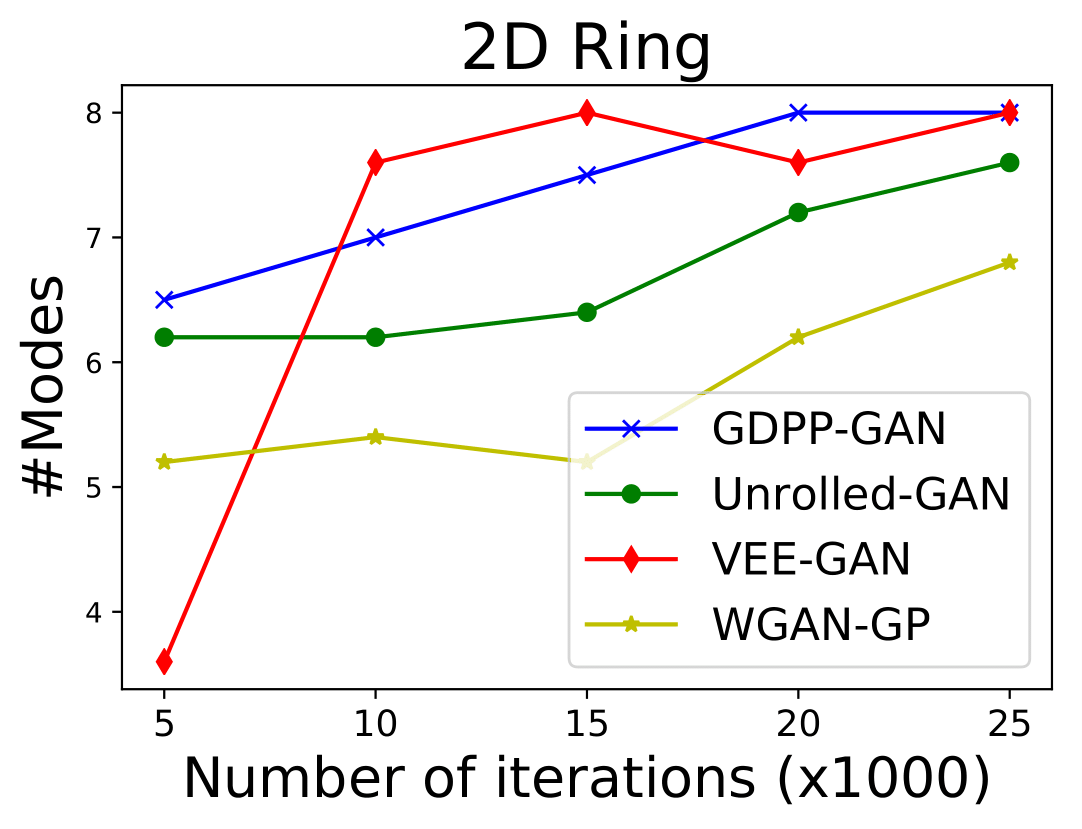}& \includegraphics[width=0.45\textwidth, height=2.5cm]{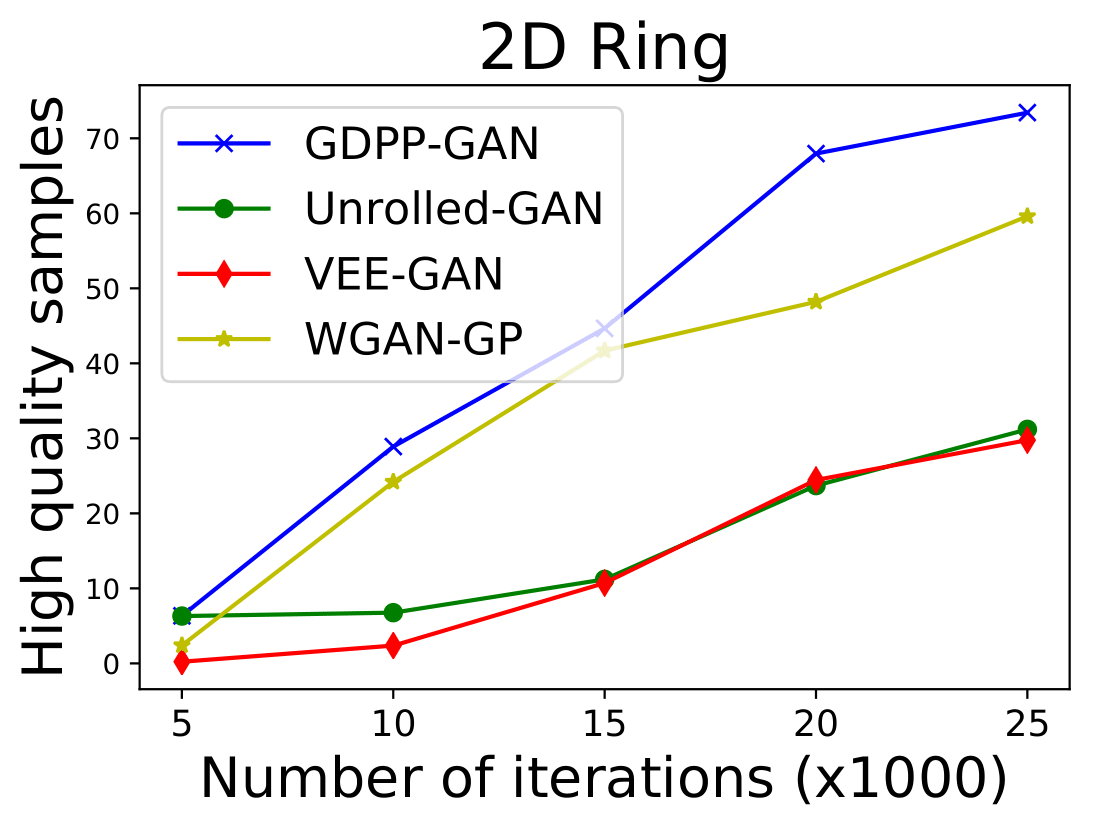}\\
    \includegraphics[width=0.45\textwidth, height=2.5cm]{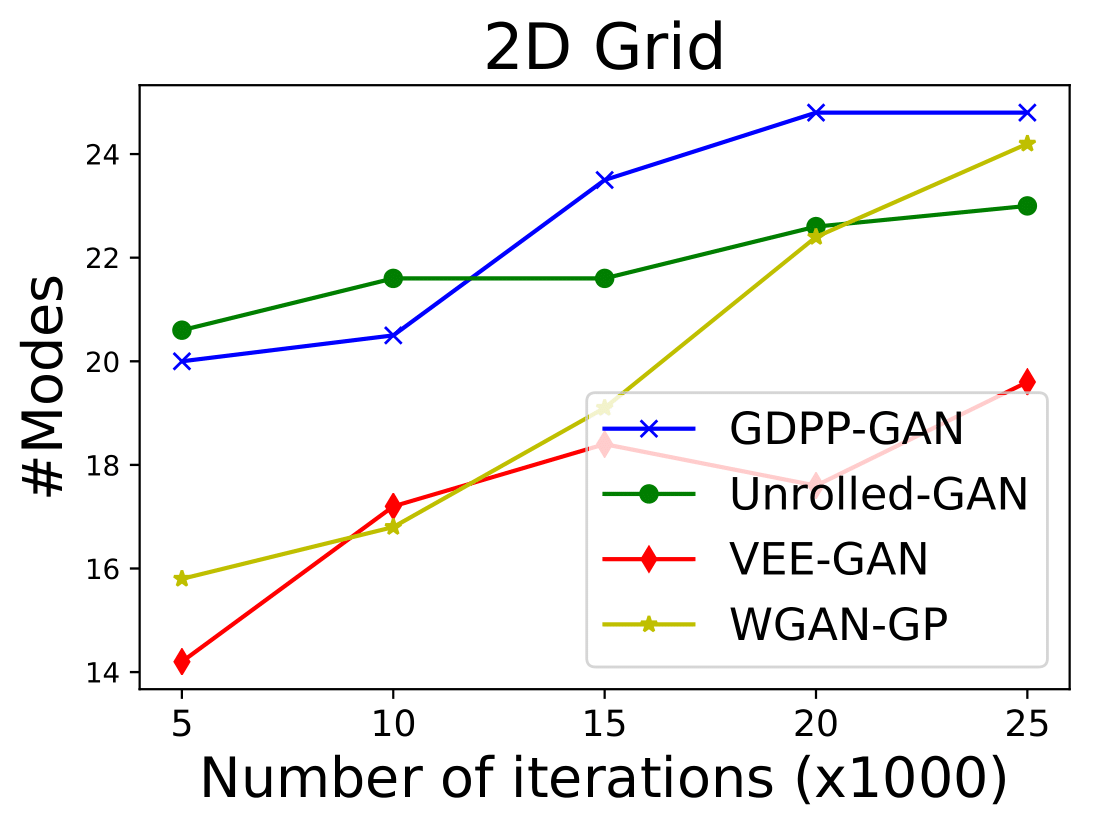}& \includegraphics[width=0.45\textwidth, height=2.5cm]{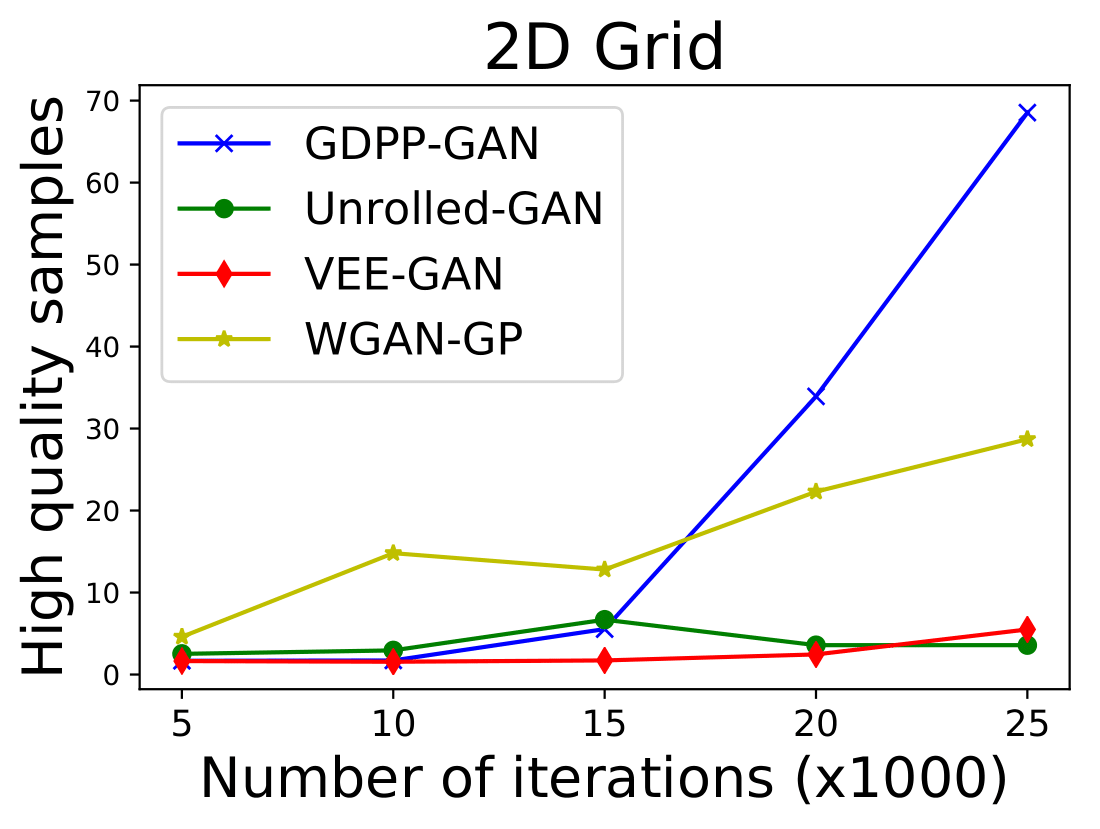}
    \end{tabular}
    \vspace{-1.5em}
    \caption{Time-Efficiency: monitoring convergence rate throughout the training given the same training data size. GDPP-GAN is the first to converge in both evaluation metrics.}
    \label{subfig:convergence_time_synth}
    \end{minipage}
    % \caption{Evaluating the models on the 2D Ring and Grid datasets in terms of (a) data-efficiency and (b) convergence-rate. GDPP-GAN tends to converge faster and require the least amount of training data.}
    \vspace{-3ex}
\end{figure*}

\textbf{Performance Evaluation:} At every iteration, we sample fake points from the generator and real points from the given distribution. Mode collapse is quantified by the number of real modes recovered in fake data, and the generation quality is quantified by the \% of High-Quality Samples. A generated sample is counted as high-quality if it was sampled within three standard deviations in case of 2D Ring or Grid, and ten standard deviations in case of the 1200D data. We train all models for 25K iterations, except for VEEGAN which needs 100K iterations to properly converge. At inference time, we generate 2500 samples from each of the trained models and measure both metrics. We report the numbers averaged over five runs with different random initialization in Table~\ref{tab:synthetic_table}. GDPP-GAN clearly outperforms all other methods, for instance on the most challenging 1200D dataset that was designed to mimic a natural data distribution, bringing a 63\% relative improvement in high-quality samples and 15\% in mode detection over its best competitor WGAN-GP. Finally, we show that our method is robust to random initialization in Appendix C.1.

% \mohamed{refer to sections in supplementary.}==> Appendix C.1
% \mohamed{\textbf{GDPP-GAN  performance:}} 

\textbf{Ablation Study:}
We run a study on the 2D Ring and Grid data to show the individual effects of each component in our loss. As shown in Table~\ref{tab:ablation}, optimizing the determinant $\det{L_S}$ directly increases the diversity generating the highest quality samples. This works best on the 2D Ring since the true data distribution can be represented by a repulsion model. However, for more complex data as in 2D Grid, optimizing the determinant fails because it does not well-represent the real manifold structure but aims at repelling the fake samples from each other. Using GDPP with an unnormalized structure term $\mathcal{L}^{u}_s$ is prone to learning outlier caused by the inherent noise within the data. Nonetheless, scaling the structure loss by the true-data eigenvalues $\hat{\lambda}$ seems to disentangle the noise from the prominent structure and better models the data diversity.

\textbf{Data-Efficiency:} We evaluate the amount of training data needed by each method to reach the same local optima as evaluated by our two metrics on both the 2D Ring and Grid data. Since the true-data is sampled from a mixture of Gaussians, we can generate an infinite size of training data. Therefore, we can quantify the amount of the training data by using the batch-size while fixing the number of back-propagation steps. In this experiment (Fig.~\ref{subfig:batch_size_synth}), we run all the methods for the same number of iterations (25,000) and vary the batch size. However, WGAN-GP tends to capture higher quality samples with fewer data. In the case of 2D Grid data, GDPP-GAN performs on par with other methods for small amounts of data, yet it tends to significantly outperform other methods on the quality of generated samples once trained on enough data.

\begin{table*}[!t]
\centering
    \begin{tabular}{lcc|cc}
    \hline
        & \multicolumn{2}{c}{Stacked-MNIST}   & \multicolumn{2}{c}{CIFAR-10}                   \\ \hline
        & \#Modes (Max 1000) & KL div. & Inception score & IvO \\
        \hline
            DCGAN \citep{radford2015unsupervised} & 427 &     3.163 & 5.26 $\pm$ 0.13 & 0.0911 \\
            DeLiGAN \citep{deligan} & 767 & 1.249 & 5.68 $\pm$ 0.09 & 0.0896 \\
            Unrolled-GAN \citep{unrolled_gan}     & 817 &     1.430 &  5.43 $\pm$ 0.21 & 0.0898 \\
            RegGAN \citep{mode_gan}               & 955 &     0.925 & 5.91 $\pm$ 0.08 & 0.0903 \\
            WGAN \citep{wgan}                     & 961 &     0.140 & 5.44 $\pm$ 0.06 & 0.0891 \\
            WGAN-GP \cite{Gulrajani2017improved}& 995 &     0.148 &  6.27 $\pm$ 0.13 & 0.0891 \\
            % \hline
             GDPP-GAN (Ours)                       & \bf 1000 & \bf 0.135 & \bf 6.58 $\pm$ 0.10 & \bf 0.0883 \\
            \hline 
            VAE \citep{vae}                     & 341 &     2.409 & 1.19 $\pm$ 0.02 & 0.543 \\
            GDPP-VAE (Ours)                     & \bf 623 &    \bf 1.328 & \bf 1.32 $\pm$ 0.03 & \bf 0.203 \\
            \hline
        \end{tabular}
  \caption{Performance of various methods on real datasets. Stacked-MNIST is evaluated using the number of captured modes (Mode Collapse) and KL-divergence between the generated class distribution and true class distribution (Quality of generations). CIFAR-10 is evaluated by Inference-via-Optimization (Mode-Collapse) and Inception-Score (Quality of generations).}
  \label{tab:cifar_results}
\end{table*}

\begin{table*}[!t]
\centering
{\small
\begin{tabular}{@{}c|c|cccccc@{}}
\toprule
 & DCGAN & Unrolled-GAN & VEE-GAN & Reg-GAN & WGAN & WGAN-GP & GDPP-GAN \\ \midrule
\begin{tabular}[c]{@{}c@{}}Avg. Iter.\\ Time (s)\end{tabular} & 0.0674 & 0.2467 & 0.1978 & 0.1357 & 0.1747 & 0.4331 & \textbf{0.0746} \\ \bottomrule
\end{tabular}
\caption{Average Iteration running time on CIFAR-10. GDPP-GAN obtains the closest time to the default (non-improved) DCGAN.}
\label{tab:running_time}}
\vspace{-1em}
\end{table*}

\textbf{Time-Efficiency:} To analyze time efficiency, we explore two primary aspects: convergence rate, and physical running time. First, to find out which method converges faster, we fix the batch size at 512 and vary the number of training iterations for all models (Fig.~\ref{subfig:convergence_time_synth}). In the 2D Ring, only VEEGAN captures a higher number of modes before GDPP-GAN, however, they are of much lower quality than the ones generated by GDPP-GAN. In 2D Grid, however, GDPP-GAN performs on par with unrolled-GAN for the first 5,000 iterations while the others are falling behind. After then, our method significantly outperforms all the methods with respect to both the number of captured modes and the quality of generated samples. Second, we compare the physical running time of all methods given the same data and number of iterations. To obtain reliable results, we chose to run the methods on CIFAR-10 instead of the synthetic, since the latter has an insignificant running time. We compute the average running time of an iteration across 1000 iterations over five different runs of each method. Table~\ref{tab:running_time} shows that GDPP-GAN has a negligible computational overhead beyond DCGAN, rendering it the fastest improved-GAN approach. We also elaborate on the run-time analysis and conduct additional experiments in Appendix C.3 to explore the computation overhead.

\subsection{Image generation experiments}
We run real-image generation experiments on three various datasets: Stacked-MNIST, CIFAR-10, and CelebA. For the first two, we use the experimental setting used in ~\citep{Gulrajani2017improved} and~\citep{unrolled_gan}. We also investigated the robustness of our method by using another more challenging setting proposed by~\citep{veegan} in Appendix C.2. For CelebA, we use the experimental setting of~\citep{karras2017progressive}. In our evaluation, we focus on comparing with the state-of-the-art methods that adopt a change in the original adversarial loss. Nevertheless, most baselines can be deemed orthogonal to our contribution and can enhance the generation if integrated with our approach. Finally, we show that our loss is generic to any generative model by incorporating it within Variational AutoEncoder (VAE) of \cite{vae} in Table ~\ref{tab:cifar_results}. Appendix D shows qualitative examples from several models and baselines.

\vspace{-1em}
\paragraph{Stacked-MNIST} A variant of MNIST~\citep{lecun1998mnist} designed to increase the number of discrete modes in the data. The data is synthesized by stacking three randomly sampled MNIST digits along the color channel resulting in a $28\times28\times3$ image. In this case, Stacked MNIST has 1000 discrete modes corresponding to the number of possible triplets of digits. Following ~\citep{Gulrajani2017improved}, we generate 50,000 images that are later used to train the networks. We train all the models for 15,000 iterations, except for DCGAN and unrolled-GAN that need 30,000 iterations to converge to a reasonable local-optima. 

We follow~\citep{veegan} to evaluate the number of recovered modes and divergence between the true and fake distributions. We sample 26000 fake images for all the models. We identify the mode of each generated image by using the classifier mentioned in \citep{mode_gan}, which is trained on the standard MNIST dataset to classify each channel of the fake sample. The quality of samples is evaluated by computing the KL-divergence between generated label distribution and training labels distribution. As shown in Table ~\ref{tab:cifar_results}, GDPP-GAN captures all modes and generates a fake distribution that has the lowest KL-Divergence with the true-distribution. Moreover, when applied on the VAE, it doubles the number of modes captured (i.e., from 341 to 623) and cuts the KL-Divergence to half (from 2.4 to 1.3). Lastly, we follow ~\citep{gans_gmms} to assess the severity of mode collapse by computing the number of statistically different bins using MNIST in Appendix C.4.

\vspace{-1em}
\paragraph{CIFAR-10} We evaluate the methods on CIFAR-10 after training all the models for 100K iterations. Unlike Stacked-MNIST, the modes are intractable in this dataset. This is why we follow ~\citep{unrolled_gan} and ~\citep{veegan} in using two different metrics: Inception Score ~\citep{improved_gan} for the generation quality and Inference-via-Optimization (IvO) for diversity. As shown in Table~\ref{tab:cifar_results}, GDPP-GAN consistently outperforms all other methods in both metrics. Furthermore, applying the GDPP on the VAE reduces the IvO by 63\%. However, we note that both the inception-scores are considerably low which is also observed by ~\cite{shmelkov2018coverage} when applying the VAE on CIFAR-10.

\begin{figure}[!t]
\centering
    \includegraphics[width=0.48\textwidth]{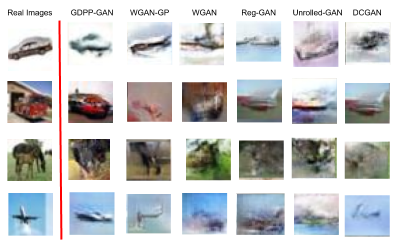}
    \vspace{-2.75em}
  \caption{Real images and their nearest generations of CIFAR-10. Nearest generations are obtained by optimizing the input noise to minimize the reconstruction error of the generated image.}
  \label{fig:nearest_neighbor}
\vspace{-1.5em}
\end{figure}

Inference-via-optimization~\citep{unrolled_gan} is used to assess the severity of mode collapse in generations by comparing real images with the nearest generated image. In the case of mode collapse, there are some real images for which this distance is large. We measure this metric by sampling a real image $x$ from the test set of real data. Then we optimize the $\ell_2$ loss between $x$ and generated image $G(z)$ by modifying the noise vector $z$. If a method attains low MSE, then it can be assumed that this method captures more modes than ones that attain a higher MSE. Fig. \ref{fig:nearest_neighbor} presents some real images with their nearest optimized generations. 

\begin{figure}[!t]
  \centering
  \includegraphics[width=0.44\textwidth, height=4.4cm]{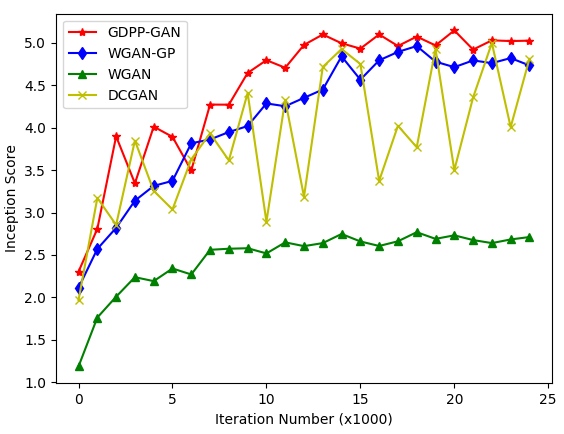}
  \vspace{-1em}
  \caption{Adding GDPP loss to DCGAN stabilizes adversarial training and generates high quality samples earliest on CIFAR-10.}
  \label{fig:cifar_incep}
  \vspace{-1em}
\end{figure}

We also assess the stability of the training, by calculating inception score at different stages while training on CIFAR-10 (Fig.~\ref{fig:cifar_incep}). Evidently, DCGAN has the least stable training with a high variation. However, by only adding GDPP penalty term to the generator loss, the model generates high-quality images the earliest on training with a stable increase.

\vspace{-1em}
\paragraph{CelebA} Finally, to evaluate the performance of our loss on large-scale adversarial training, we embed our GDPP loss in Progressive-Growing GANs~\citep{karras2017progressive}. We train the models for 40K iterations corresponding to 4 scales up to $64\times64$ results, and for 200K iterations at 5 scales ($128\times128$). On large scale datasets such as CelebA dataset~\citep{celebA}, it is harder to stabilize the training of DCGAN. In fact, DCGAN is only able to produce reasonable results in the first scale but not the second due to the high-resolution requirement. That is why, we embed our loss with WGAN-GP this time instead of DCGAN paradigm, which is as well orthogonal to our loss. 

Unlike CIFAR-10 dataset, CelebA does not simulate ImageNet because it only contains faces, not natural scenes/objects. Therefore, using a model trained on ImageNet as a basis for evaluation (i.e., Inception Score), will cause inaccurate recognition. On the other hand, IvO was shown to be fooled by producing blurry images out of the optimization in high-resolution datasets as in CelebA~\citep{veegan}. Therefore, we follow ~\cite{karras2017progressive} to evaluate the performance on CelebA using Sliced Wasserstein Distance (SWD)~\citep{swd}. A small Wasserstein distance indicates that the distribution of the patches is similar, which entails that real and fake images appear similar in both appearance and variation at this spatial resolution. Accordingly, the SWD metric can evaluate the quality of images as well as the severity of mode-collapse on large-scale datasets such as CelebA. Table~\ref{tab:swd} shows the average and minimum SWD metric across the last 10K training iterations. We chose this time frame because it shows a saturation in training loss for all the competing methods.

\begin{table}[!t]
    \centering
    \begin{tabular}{cccc}
\hline
&& Avg. SWD & Min. SWD  \\ \hline
\parbox[t]{2mm}{\multirow{4}{*}{\rotatebox[origin=l]{90}{$~~64\times64~~$}}}   &Training Data & \multicolumn{2}{c}{0.0033} \\ \cline{2-4} 
& DCGAN &  0.0906 & 0.0241  \\  
                             & WGAN-GP                      & 0.0186 & 0.0115  \\  
                             &  GDPP-GAN                                             & \textbf{0.0163} & \textbf{0.0075} \\ 
                             \hline
\parbox[t]{2mm}{\multirow{4}{*}{\rotatebox[origin=l]{90}{$~~~128^2$}}}   
&Training Data & \multicolumn{2}{c}{0.0023} \\ \cline{2-4} 
&WGAN-GP & 0.0197  & 0.0095 \\
&GDPP-GAN & \textbf{0.0181} & \textbf{0.0088} \\
\hline
\end{tabular}
\caption{Average and Minimum Sliced Wasserstein Distance over the last 10K iterations at scales $64^2$, and scales $128^2$ on CelebA.  Training Data is the upper limit for this metric.}
\label{tab:swd}
\vspace{-3ex}
\end{table}

\section{Conclusion}
In this work, we introduced a novel criterion to train generative networks on capturing a similar diversity to one of the true data by utilizing Determinantal Point Process(DPP). We apply our criterion to Generative Adversarial training and the Variational AutoEncoder by learning a kernel via features extracted from the discriminator/encoder. Then, we train the generator on optimizing a loss between the fake and real, eigenvalues and eigenvectors of this kernel to encourage the generator on simulating the diversity of real data. Our GDPP framework accumulates many desirable properties: it does not require any extra trainable parameters, it operates in an unsupervised setting, yet it consistently outperforms state-of-the-art methods on a battery of synthetic data and real image datasets as measure by generation quality and invariance to mode collapse. Furthermore, GDPP-GANs exhibit a stabilized adversarial training and has been shown to be time and data efficient as compared to state-of-the-art approaches. Moreover, the GDPP criterion is architecture and model invariant, allowing it to be embedded with any variants of generative models such as adversarial feature learning and conditional GANs.

%Carreful, 9 pages again! Camille going to bed.
% Will handle it... Have a great night... Thanks a lot for your help ^_^ 

% Therefore, we aim at future work to explore incorporating the GDPP criterion with orthogonal methods such as adversarial feature learning, conditional GANs, as well as other generative models. 

%Higher Quality Images. state-of-the-art Inception-score on CIFAR-10\\
 % can't provide the url unless anonymized 
%Code available at \url{https://github.com/M-Elfeki/DPP-GAN}.

%\mohamed{ However, we think that our GDPP loss can be simply  integrated with other variations of generative models.}
 
% 8 pages, plus references

\nocite{langley00}
\bibliography{example_paper}
\bibliographystyle{icml2019}

\end{document}